\documentclass[11pt,a4paper]{article}
\usepackage[hyperref]{acl2021}
\usepackage{times}
\usepackage{latexsym}

\usepackage{booktabs}
\usepackage{enumitem}
\usepackage{dirtytalk}
\usepackage{microtype}
\usepackage{graphicx}

\aclfinalcopy %

\newcommand{\posscite}[1]{\citeauthor{#1}'s (\citeyear{#1})}

\title{Underreporting of errors in NLG output, and what to do about it}

\author{
\parbox{\textwidth}{\centering 
Emiel~van~Miltenburg,$^{1}$\thanks{\hphantom{$^{*}$}This project was led by the first author. Remaining authors are presented in alphabetical order.}\hspace{1ex}
Miruna~Clinciu,$^{2,3,4}$ 
Ond\v{r}ej~Du\v{s}ek,$^{5}$ 
Dimitra~Gkatzia,$^{6}$  
Stephanie~Inglis,$^{7}$  
Leo~Leppänen,$^{8}$  
Saad~Mahamood,$^{9}$  
Emma~Manning,$^{10}$  
Stephanie~Schoch,$^{11}$  
Craig~Thomson,$^{7}$  {\normalfont and} 
Luou~Wen$^{12}$}\\[3ex]  
\parbox{\textwidth}{\centering 
$^{1}$Tilburg~University,
$^{2}$Edinburgh~Centre~for~Robotics,
$^{3}$Heriot-Watt~University,
$^{4}$University~of~Edinburgh,
$^{5}$Charles~University,~Prague,
$^{6}$Edinburgh~Napier~University, 
$^{7}$University~of~Aberdeen, 
$^{8}$University~of~Helsinki, 
$^{9}$trivago~N.V., 
$^{10}$Georgetown~University, 
$^{11}$University~of~Virginia,
$^{12}$University~of~Cambridge} \\
Contact: \texttt{c.w.j.vanmiltenburg@tilburguniversity.edu} \\}

\date{}

\begin{document}
\maketitle
\begin{abstract}
We observe a severe under-reporting of the different kinds of errors that Natural Language Generation systems make. This is a problem, because mistakes are an important indicator of where systems should still be improved. If authors only report overall performance metrics, the research community is left in the dark about the specific weaknesses that are exhibited by `state-of-the-art' research. Next to quantifying the extent of error under-reporting, this position paper provides recommendations for error identification, analysis and reporting.
\end{abstract}

\section{Introduction}
This paper turned out very differently from the one we had initially intended to write. Our original intention was to write an overview of the different kinds of errors that appear in the output of different kinds of Natural Language Generation (NLG) systems, and to develop a general taxonomy of NLG output errors, based on the publications that have appeared at previous INLG conferences (similar to \citealt{howcroft-etal-2020-twenty,belz-etal-2020-disentangling}). This, however, turned out to be impossible. The reason? There is a severe under-reporting of the different kinds of errors that NLG systems make. By this assertion, we mean that authors neither include any error analysis nor provide any examples of errors made by the system, and they do not make reference to different kinds of errors that may appear in the output. The latter is a lower bar than carrying out an error analysis, which requires a more systematic approach where several outputs are sampled and analysed for the presence of errors, which are then categorised (ideally through a formal procedure with multiple annotators). Section~\ref{sec:underreporting} provides more detailed statistics about error reporting in different years of INLG (and ENLG), and the amount of papers that discuss the kinds of errors that may appear in NLG output.

The fact that errors are under-reported in the NLG literature is probably unsurprising to experienced researchers in this area. The lack of reporting of negative results in AI has been a well-known issue for many years \cite{REITER200341}. With the classic NLG example being the reporting of negative results for the STOP project on smoking cessation \cite{reiter-etal-2001-using, REITER200341}. But even going in with (relatively) low expectations, it was confronting just to see how little we as a community look at the mistakes that our systems make.

We believe that it is both necessary \emph{and} possible to improve our ways. One of the reasons why it is necessary to provide more error analyses (see \S\ref{sec:whyreport} for more), is that otherwise, it is unclear what are the strengths and weaknesses of current NLG systems. In what follows, we provide guidance on how to gain more insight into system behavior.

This paper provides a general framework to carry out error analyses. First we cover the terminology and related literature (\S\ref{sec:background}), after which we quantify the problem of under-reporting (\S\ref{sec:underreporting}). Following up on this, we provide recommendations on how to carry out an error analysis (\S\ref{sec:recommendations}). We acknowledge that there are barriers to a more widespread adoption of error analyses, and discuss some ways to overcome them (\S\ref{sec:future}). Our code and data are provided as supplementary materials.

\section{Background: NLG systems and errors}\label{sec:background}%

\subsection{Defining errors}
There are many ways in which a given NLG system can fail. Therefore it can be difficult to exactly define all the different types of errors that can possibly occur. Whilst error analyses in past NLG literature were not sufficient for us to create a taxonomy, we will instead propose high-level distinctions to help bring clarity within the NLG research community.

This paper focuses on text errors, that we define as countable instances of things that went wrong, as identified from the generated text.\footnote{We use the term `text' to refer to any expression of natural language. For example, sign language (as in \citealt{mazzei-2015-translating}) would be considered `text' under this definition.}  We focus on text errors, where something is incorrect in the generated text with respect to the data, an external knowledge source, or the communicative goal. %

Through our focus on text errors, we only look at the \emph{product} (what comes out) of an NLG system, so that we can compare the result of different kinds of systems (e.g., rule-based pipelines versus neural end-to-end systems), with error categories that are independent of the \emph{process} (how the text is produced).\footnote{By focusing on countable instances of things that went wrong in the output text, we also exclude issues such as bias and low output diversity, that are global properties of the collection of outputs that a system produces for a given amount of inputs, rather than being identifiable in individual outputs.} For completeness, we discuss errors related to the production process in Section~\ref{sec:process-level}.

By \emph{error analysis} we mean the identification and categorisation of errors, after which statistics about the distribution of error categories are reported. It is an annotation process \cite{pustejovsky2012natural,ide2017handbook}, similar to Quantitative Content Analysis in the social sciences \cite{krippendorff2018content,neuendorf2017content}.\footnote{There has been some effort to automate this process. For example, \citet{shimorina2021error} describe an automatic error analysis procedure for shallow surface realisation, and \citet{stevens-guille-etal-2020-neural} automate the detection of repetitions, omissions, and hallucinations. However, for many NLG tasks, this kind of automation is still out of reach, given the wide range of possible correct outputs that are available in language generation tasks.} 
Error analysis can be carried out during development (to see what kinds of mistakes the system is currently making), as the last part of a study (evaluating a new system that you are presenting), or as a standalone study (comparing different systems). The latter option requires output data to be available, ideally for both the validation and test sets. A rich source of output data is the GEM shared task \cite{gehrmann2021gem}. 

Text errors can be categorised in several different types, including factual errors (e.g. incorrect number;  \citealt{thomson-reiter-2020-gold}), and errors related to form (spelling, grammaticality), style (formal versus informal, empathetic versus neutral), or behavior (over- and under-specification). Some of these are universally wrong, while others may be `contextually wrong' with respect to the task success or for a particular design goal. For example, formal texts aren't wrong \emph{per se}, but if the goal is to produce informal texts, then any semblance of formality may be considered incorrect. 

It may be possible to relate different kinds of errors to the different dimensions of text quality identified by \citet{belz-etal-2020-disentangling}. What is crucial here, is that we are able to identify the specific thing which went wrong, rather than just generate a number that is representative of overall quality.

\subsection{Why do authors need to report errors?}\label{sec:whyreport}
There is a need for realism in the NLG community. By providing examples of different kinds of errors, we can show the complexity of the task(s) at hand, and the challenges that still lie ahead. This also helps set realistic expectations for users of NLG technology, and people who might otherwise build on top of our work. %
A similar argument has been put forward by \citet{mitchell2019model}, arguing for `model cards' that provide, inter alia, performance metrics based on quantitative evaluation methods. We encourage authors to also look at the data and provide examples of where systems produce errors. Under-reporting the types of errors that a system makes is harmful because it leaves us unable to fully appreciate the system's performance.

While some errors may be detected automatically, e.g., using information extraction techniques \citep{wiseman-etal-2017-challenges} or manually defined rules \citep{dusek-etal-2018-findings},  others are harder or impossible to identify if not reported. We rely on researchers to communicate the less obvious errors to the reader, to avoid them going unnoticed and causing harm for subsequent users of the technology.

Reporting errors is also useful when comparing different implementation paradigms, such as pipeline-based data-to-text systems versus neural end-to-end systems. It is important to ask where systems fall short, because different systems may have different shortcomings. One example of this is the E2E challenge, where systems with similar human rating scores show very different behavior \cite{dusek_evaluating_2020}.

Finally, human and automatic evaluation metrics, or at least the ones that generate some kind of intrinsic rating, are too coarse-grained to capture relevant information. They are general evaluations of system performance that estimate an average-case performance across a limited set of abstract dimensions (if they measure anything meaningful at all; see \citealt{reiter-2018-structured}). We don't usually know the worst-case performance, and we don't know what kinds of errors cause the metrics or ratings to be sub-optimal. Additionally, the general lack of extrinsic evaluations among NLG researchers \cite{gkatzia2015snapshot} means that in some cases we only have a partial understanding of the possible errors for a given system.

\subsection{Levels of analysis}\label{sec:process-level}
As noted above, our focus on errors in the output text is essential to facilitate framework-neutral comparisons between the performance of different systems. When categorizing the errors made by different systems, it is important to be careful with terms such as \emph{hallucination} and \emph{omission}, since these are process-level (pertaining to the system) rather than product-level (pertaining to the output) descriptions of the errors.\footnote{Furthermore, terms like \emph{hallucination} may be seen as unnecessary anthropomorphisms, that trivialise mental illness.} Process-level descriptions are problematic because we cannot reliably determine how an error came about, based on the output alone.\footnote{A further reason to avoid process-level descriptors is that they are often strongly associated with one type of approach. For example, the term `hallucination' is almost exclusively used with end-to-end systems, as it is common for these systems to add phrases in the output text that are not grounded in the input. In our experience, pipeline systems are hardly ever referred to as `hallucinating.' As such, it is better to avoid the term and instead talk about concrete phenomena in the output.} We can distinguish between at least two causes of errors, that we define below: system problems and data problems. While these problems should be dealt with, we do not consider these to be the subject of error analysis.

\textbf{System problems} can be defined as the malfunctioning of one or several components in a given system, or the malfunctioning of the system as a whole. System problems in rule/template-based systems could be considered as synonymous to `bugs,' which are either semantic and/or syntactic in nature. If the system has operated in a mode other than intended (e.g., as spotted through an error analysis), the problem has to be identified, and then corrected. Identifying and solving such problems may require close involvement of domain experts for systems that incorporate significant domain knowledge or expertise \cite{mahamood2012working}. \Citet{deemter2018lying} provide further discussion of how errors could occur at different stages of the NLG pipeline system. System problems in end-to-end systems are harder to identify, but recent work on interpretability/explainability aims to improve this \cite{Gilpin2019}.

\textbf{Data problems} are inaccuracies in the input that are reflected in the output. For example: when a player scored three goals in a real-world sports game, but only one goal is recorded (for whatever reason) in the data, even a perfect NLG system will generate an error in its summary of the match. Such errors may be identified as factual errors by cross-referencing the input data with external sources. They can then be further diagnosed as data errors by tracing back the errors to the data source.

\begin{table*}
    \small\centering
\begin{tabular}{lccccc}
\toprule
 Venue    &   Total &   Amenable &   Error mention &   Error analysis &   Percentage with error analysis \\
\midrule
 INLG2010 &      37 &         16 &               \hphantom{0}6 &                0 &                        \hphantom{0}0\% \\
 ENLG2015 &      28 &         20 &               \hphantom{0}4 &                1 &                       \hphantom{0}5\% \\
 INLG2020 &      46 &         35 &              19 &                4 &                       11\% \\
\bottomrule
\end{tabular}
    \caption{Annotation results for different SIGGEN conferences, showing the percentage of amenable papers that included error analyses. Upon further inspection, most error mentions are relatively general/superficial.}
    \label{tab:annotationresults}
\end{table*}

\section{Under-reporting of errors}\label{sec:underreporting}
We examined different *NLG conferences to determine the amount of papers that describe (types of) output errors, and the amount of papers that actually provide a manual error analysis.

\subsection{Approach}
We selected all the papers from three SIGGEN conferences, five years apart from each other: \textsc{inlg2010, enlg2015}, and \textsc{inlg2020}. We split up the papers such that all authors looked at a selection of papers from one of these conferences, and informally marked all papers that discuss NLG errors in some way. These papers helped us define the terms `error' and `error analysis' more precisely.

In a second round of annotation, multiple annotators categorised all papers as `amenable' or `not amenable' to an error analysis. A paper is amenable to an error analysis if one of its primary contributions is presenting an NLG system that produces some form of output text. So, NLG experiments are amenable to an error analysis, while survey papers are not.\footnote{Examples of other kinds of papers that are not amenable include evaluation papers, shared task proposals, papers which analyze patterns in human-produced language, and papers which describe a component in ongoing NLG work which does not yet produce textual output (e.g.~a ranking module).} For all amenable papers, the annotator indicated whether the paper (a) mentions any errors in the output and (b) whether it contains an error analysis.\footnote{As defined in Section~\ref{sec:background}, errors are (countable) instances of something that is wrong about the output. An `error mention' is a reference to such an instance or a class of such instances. Error analyses are formalised procedures through which annotators identify and categorise errors in the output.} We encouraged discussion between annotators whenever they felt uncertain (details in Appendix~\ref{app:annotation}). The annotations for each paper were subsequently checked by one other annotator, after which any disagreements were adjudicated through a group discussion.

\subsection{Results}
Table~\ref{tab:annotationresults} provides an overview of our results. We found that only five papers at the selected *NLG conferences provide an error analysis,\footnote{Summaries of these error analyses are in Appendix~\ref{app:summaries}.} and more than half of the papers fail to mention any errors in the output. %
This means that the INLG community is systematically under-informed about the weaknesses of existing approaches. In light of our original goal, it does not seem to be a fruitful exercise to survey all SIGGEN papers if so few authors discuss any output errors. Instead, we need a culture change where authors discuss the output of their systems in more detail. Once this practice is more common, we can start to make generalisations about the different kinds of errors that NLG systems make. To facilitate this culture change, we give a set of recommendations for error analysis.

\section{Recommendations for error analysis}
\label{sec:recommendations}

\begin{figure*}
\centering
\includegraphics[width=0.9\textwidth]{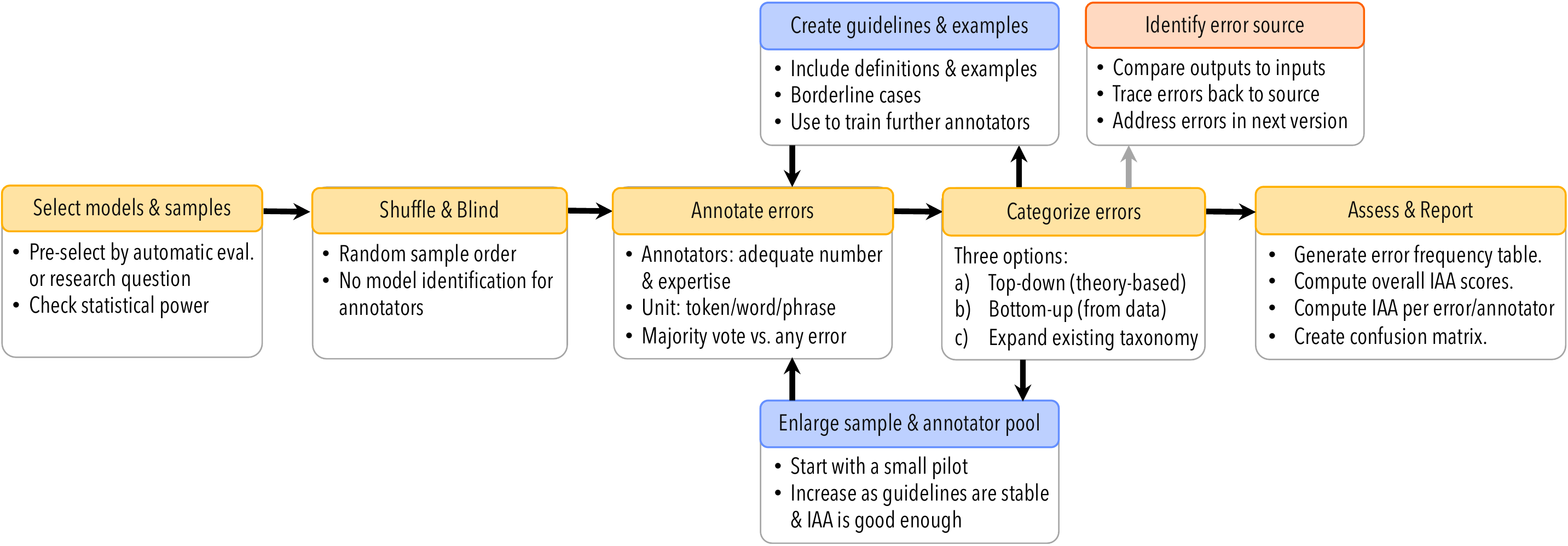}
\caption{Flowchart depicting recommended analysis steps, as described in Section~\ref{sec:recommendations}. IAA stands for Inter-Annotator Agreement, as measured through Cohen's kappa or Krippendorff's alpha, for example.}
\label{fig:flowchart}
\end{figure*}

We provide general recommendations for carrying out an error analysis, summarized in Figure~\ref{fig:flowchart}.

\subsection{Setting expectations}
\label{sec:setting-expectations}
Before starting, it is important to be clear about your goals and expectations for the study.

\paragraph{Goal}
Generally speaking, the goal of an error analysis is to find and quantify system errors statistically, to allow a thorough comparison of different systems, and to help the reader understand the shortcomings of your system. But your personal goals and interests may differ. For example, you may only be interested in \emph{grammatical} errors, and less so in factual errors.

\paragraph{Expected errors}
When starting an error analysis, you may already have some ideas about what kinds of errors might appear in the outputs of different systems. These ideas may stem from the literature (theoretical limitations, or discussions of errors), from your personal experience as an NLG researcher, or it might just be an impression you have from talking to others. You might also have particular expectations about what the distribution of errors will look like.

Both goals and expectations may bias your study, and cause you to overlook particular kinds of errors. But if you are aware of these biases, you may be able to take them into account, and later check if the results confirm your original expectations. Hence, it may be useful to preregister your study, so as to make your thoughts and plans explicit \cite{doi:10.1080/08989621.2019.1580147,van-miltenburg-etal-2021-preregistering}. This also makes it easier for others to check whether they agree with the assumptions behind your study.

\subsection{Set-up}
\label{sec:analysis-setup}
Given your goals and expectations, there are several design choices that you have to make, in order to carry out your study.

\paragraph{Systems and outputs} Since error analysis is relatively labor-intensive, it may not be feasible to look at a wide array of different systems. In that case, you could pre-select a smaller number of models, either based on automatic metric scores, or based on specific model features you are interested in. Alternatively, you could see to what extent the model outputs overlap, given the same input. If two models produce exactly the same output, you only need to annotate that output once.

\paragraph{Number of outputs}
Ideally, the number of outputs should be based on a power analysis to provide meaningful comparisons \cite{card-etal-2020-little,van_der_lee_human_2021}, but other considerations, such as time and budget, may be taken into account.

\paragraph{Sample selection} 
Regarding the selection of examples to analyze, there are three basic alternatives: The most basic is \emph{random sampling} from the validation/test outputs. Another option is selecting \emph{specific kinds of inputs} and analysing all corresponding outputs. Here, inputs known to be difficult/adversarial or inputs specifically targeting system properties or features may be selected \cite{ribeiro-etal-2020-beyond}. Finally, examples to analyze may also be selected based on \emph{quantitative values}: automatic metric scores or ratings in a preceding general human evaluation. This way, error analysis can provide explanation for the automatic or human scores.
The most suitable option depends on your specific use case: While random selection gives the least biased picture of the model performance, selecting specifically hard and/or low-rated samples may be more efficient. Also note that the sample selection should always be independent of any samples you may have previously examined during system development, since any errors for those cases are likely to have been resolved already (although you cannot be sure until you have verified these cases as well). 

\paragraph{Presentation} The order of the items should be randomized (to reduce possible order effects), and if multiple system variants are considered, the annotators must not know which system produced which output (to minimise annotator bias).

\paragraph{Interface} The efficiency of any annotation task depends on the quality of the interface. With the right interface, annotators may be able to annotate more data in a shorter time frame. \citet[Chapter 11]{monarch2021human} provides recommendations on interface design based on principles from the field of Human-Computer Interaction (HCI). Once you have a working interface, it is important to test the interface and obtain feedback from annotators to see whether it can be made more intuitive or efficient (e.g.\ by adding keyboard shortcuts to perform common operations. Also note that keyboard operations are generally quicker than using the mouse).

\paragraph {Annotators and the annotation process}
The annotation process can be split into two parts: identifying the errors (\S\ref{sec:identifying-errors}), and categorising the errors (\S\ref{sec:categorization-system}). These can either be carried out sequentially (first identify, then categorize) or simultaneously (asking annotators to both identify and categorize errors at the same time). The choices you make here also impact annotator requirements, and the evaluation of the annotation procedure.

\paragraph{Number of annotators} Generally speaking, having more annotators reduces the prevalence of the individual bias \cite{Artstein2005}. This is particularly relevant if we want to detect all the errors in the output data. Having more annotators means that we are less likely to overlook individual instances of errors. Once those errors are identified, it may make more sense to rely on a smaller set of well-trained annotators to categorise the different errors. In the ideal situation, all errors are annotated by (at least) two judges so as to be able to detect and resolve any disagreements afterwards. If this is not possible, then you should at least double-annotate a large enough sample to reliably estimate inter-annotator agreement.\footnote{See \citealt{krippendorff2011agreement} for a reference table to determine the sample size for Krippendorff's $\alpha$. Similar studies exist for Cohen's $\kappa$, e.g. \citealt{flack1988sample,10.1093/ptj/85.3.257}.}

\paragraph{Role of the annotators} Ideally, the annotators should be independent of the authors reporting the error analysis \cite{neuendorf2017content}, to ensure that the results are not influenced by any personal biases about the systems involved, and that the annotations are indeed based on the guidelines themselves rather than on discussions between the authors. If this is not feasible, then the authors should at least ensure that they remain ignorant of the identity of the system that produced the relevant outputs.

\paragraph{Degree of expertise}
Depending on the complexity of the annotation guidelines, the error analysis may require expertise in linguistics (in the case of a theory-driven error categorisation scheme), or the relevant application area (with a context-driven error categorisation scheme). For example, \citet{mahamood2012working} worked with nurses to identify errors in reports generated for parents of neonatal infants. Taking into consideration the costly process of selecting domain expert annotators and the importance of quality control, non-domain experts might be also considered, ensuring their qualification through (intensive) training \cite{Artstein2005, Carlson2001}.\footnote{At least on the MTurk platform, Requesters can set the entrance requirements for their tasks such that only Workers who passed a qualifying test may carry out annotation tasks.}

\paragraph{Compensation and treatment of workers}
If annotators are hired, either directly or via a crowd-sourcing platform such as MTurk, they should be compensated and treated fairly \cite{fort-etal-2011-last}. \citet{Silberman-responsible-2018} provide useful guidelines for the treatment of crowd-workers. The authors note that they should at least be paid the minimum wage, they should be paid promptly, and they should be treated with respect. This means you should be ready to answer questions about the annotation task, and to streamline the task based on worker feedback. If you use human participants to annotate the data, you likely also need to apply for approval by an Institutional Review Board (IRB).

\paragraph{Training}
Annotators should receive training to be able to carry out the error analysis, but the amount of training depends on the difficulty of the task (which depends, among other factors, on the \emph{coding units} (see \S\ref{sec:identifying-errors}), and the number of error types to distinguish). %
They should be provided with the annotation guidelines (\S\ref{sec:guidelines}), and then be asked to annotate texts where the errors are known (but not visible). The solutions would ideally be created by experts, although in some cases, solutions created by researchers may be sufficient \citep{thomson-reiter-2020-gold}.  It should be decided in advance what the threshold is to accept annotators for the reaming work, and, if they fail, whether to provide additional training or find other candidates. Note that annotators should also be compensated for taking part in the training (see previous paragraph).

\subsection{Identifying the errors}
\label{sec:identifying-errors}

Error identification focuses on discovering all errors in the chosen output samples (as defined in the introduction). Previously, \citet{popovic-2020-informative} asked error annotators to identify issues with comprehensibility and adequacy in machine-translated text. Similarly, \citet{freitag2021experts} proposed a manual error annotation task where the annotators identified and highlighted errors within each segment in a document, taking into account the document's context as well as the severity of the errors.

The major challenge in this annotation step is how to determine the units of analysis; should annotators mark individual tokens, phrases, or constituents as being incorrect, or can they just freely highlight any sequence of words? In content analysis, this is called \emph{unitizing}, and having an agreed-upon unit of analysis makes it easier to process the annotations and compute inter-annotator agreement \cite{krippendorff2016reliability}.\footnote{Though note that \citeauthor{krippendorff2016reliability} do provide a metric to compute inter-annotator agreement for annotators who use units of different lengths.} What is the right unit may depend on the task at hand, and as such is beyond the scope of this paper.\footnote{One interesting solution to the problem of unitization is provided by \citet{pagnoni-etal-2021-understanding}, who do not identify individual errors, but do allow annotators to ``check all types that apply'' at the sentence level. The downside of this approach is that it is not fine-grained enough to be able to count individual instances of errors, but you do get an overall impression of the error distribution based on the sentence count for each type.}

A final question is what to do when there is disagreement between annotators about what counts as an error or not. When working with multiple annotators, it may be possible to use majority voting, but one might also be inclusive and keep all the identified errors for further annotation. The error categorization phase may then include a category for those instances that are not errors after all.

\subsection{Categorizing errors}
\label{sec:categorization-system}

There are three ways to develop an error categorisation system:
\begin{enumerate}[itemsep=5px,wide,topsep=0px]
    \item \textbf{Top-down} approaches use existing theory to derive different types of errors. For example, \citet{higashinaka-etal-2015-towards} develop an error taxonomy based on \posscite{LogicandConversation} Maxims of conversation. And the top levels of \posscite{costa2015linguistically} error taxonomy\footnote{Orthography, Lexis, Grammar, Semantic, and Discourse.} are based on general linguistic theory, inspired by \citet{dulay1982language}.
    \item \textbf{Bottom-up} approaches first identify different errors in the output, and then try to develop coherent categories of errors based on the different kinds of attested errors. An example of this is provided by \citet{higashinaka-etal-2015-fatal}, who use a clustering algorithm to automatically group errors based on comments from the annotators (verbal descriptions of the nature of the mistakes that were made). Of course, you do not have to use a clustering algorithm. You can also manually sort the errors into different groups (either digitally\footnote{E.g.\ via a program like Excel, \href{https://www.maxqda.com}{MaxQDA} or \href{https://atlasti.com}{Atlas.ti}, or a website like \url{https://www.well-sorted.org}.} or physically\footnote{A good example of this \emph{pile sorting} method is provided by \citet{yeh2014sorting}. \citet{blanchard2016evidence} give further recommendations.}).
    \item \textbf{Expanding on existing taxonomies}: here we make use of other researchers' efforts to categorize different kinds of errors, by adding, removing, or merging different categories. For example, \citet{costa2015linguistically} describe how different taxonomies of errors in Machine Translation build on each other. In NLG, if you are working on data-to-text, then you could take \posscite{thomson-reiter-2020-gold} taxonomy as a starting point. Alternatively, \citet{dou2021scarecrow} present a crowd-sourced error annotation schema called \textsc{scarecrow}. For image captioning, there is a more specific taxonomy provided by \citet{DBLP:journals/corr/MiltenburgE17}. Future work may also investigate the possibility of merging all of these taxonomies and relating the categories to the quality criteria identified by \citet{belz-etal-2020-disentangling}.
\end{enumerate}

\paragraph{The problem of error ambiguity} To be able to categorize different kinds of errors, we often rely on the \emph{edit-distance heuristic}. That is: we say that the text contains particular kinds of errors, because fixing those errors will give us the desired output. With this reasoning, we take the mental `shortest path' towards the closest correct text.\footnote{Note that we don't know whether the errors we identified are actually the ones that the system internally made. This would require further investigation, tracing back the origins of each different instance of an error.} This at least gives us a set of `perceived errors' in the text, that provides a useful starting point for future research. However, during the process of identifying errors, we may find that there are multiple `shortest paths' that lead to a correct utterance, resulting in error ambiguity (see, e.g., \Citealt{DBLP:journals/corr/MiltenburgE17}; \citealt[\S 3.3]{thomson-reiter-2020-gold}). 

For example, if the output text from a sports summary system notes that Player A scored 2 points, while in fact Player A scored 1 point and Player B scored 2 points, should we say that this is a number error (2 instead of 1) or a person error (Player A instead of B)? This example also shows the fragility of the distinction between product and process. It is very tempting to look at what the system did to determine the right category, but it is unclear whether the `true error category' is always knowable.

There are multiple ways to address the problem of error ambiguity. For instance, we may award partial credit ($1/n$ error categories), mark both types of errors as applying in this situation (overgeneralising, to be on the safe side), or count all ambiguous cases to separately report on them in the overall frequency table. Another solution, used by \citet{thomson-reiter-2020-gold} is to provide the annotators with a fixed preference order (\textsc{name, number, word, context}), so that similar cases are resolved in a similar fashion. 

\subsection{Writing annotation guidelines}\label{sec:guidelines}
Once you have determined an error identification strategy and developed an error categorisation system, you should describe these in a clear set of annotation guidelines. At the very least, these guidelines should contain relevant definitions (of each error category, and of errors in general), along with a set of examples, so that annotators can easily recognize different types of errors. For clarity, you may wish to add examples of borderline cases with an explanation of why they should be categorized in a particular way.

\paragraph{Pilot} The development of a categorisation system and matching guidelines is an iterative process. This means that you will need to carry out multiple pilot studies in order to end up with a reliable set of guidelines,\footnote{As determined by an inter-annotator agreement that exceeds a particular threshold, e.g. Krippendorff's $\alpha\geq0.8$.} that is easily understood by the annotators, and provides full coverage of the data. Pilot studies are also important to determine how long the annotation will take. This is not just practical to plan your study, but also essential to determine how much crowd-workers should be paid per task, so that you are able to guarantee a minimum wage.

\subsection{Assessment}
Annotators and annotations can be assessed during or after the error analysis.\footnote{And in many cases, the annotators will already have been assessed during the training phase, using the same measures.}

\paragraph{During the error analysis} Particularly with crowd-sourced annotations it is common to include gold-standard items in the annotation task, so that it is possible to flag annotators who provide too many incorrect responses. It is also possible to carry out an intermediate assessment of inter-annotator agreement (IAA), described in more detail below. This is particularly relevant for larger projects, where annotators may diverge over time.

\paragraph{After the error analysis} You can compute IAA scores (e.g., Cohen's $\kappa$ or Krippendorff's $\alpha$, see: \citealt{cohen1960coefficient,krippendorff1970bivariate,krippendorff2018content}), to show the overall reliability of the annotations, the pairwise agreement between different annotators, and the reliability of the annotations for each error type. You can also produce a confusion matrix; a table that takes one of the annotators (or the adjudicated annotations after discussion) as a reference, and provides counts for how often errors from a particular category were annotated as belonging to any of the error categories \citep{pustejovsky2012natural}. This shows all disagreements at a glance.

Any analysis of (dis)agreement or IAA scores requires there to be overlap between the annotators. This overlap should be large enough to reliably identify any issues with either the guidelines or the annotators.
Low agreement between annotators may be addressed by having an adjudication round, where the annotators (or an expert judge) resolve any disagreements; rejecting the work of unreliable annotators; or revising the task or the annotation guidelines, followed by another annotation round \citep{pustejovsky2012natural}.

\subsection{Reporting}
We recommend that authors should provide a table reporting the frequency of each error type, along with the relevant IAA scores. The main text should at least provide the overall IAA score, while IAA scores for the separate error categories could also be provided in the appendix. For completeness, it is also useful to include a confusion matrix, but this can also be put in the appendix. The main text should provide a discussion of both the frequency table, as well as the IAA scores. What might explain the distribution of errors? What do the examples from the \emph{Other}-category look like? And how should we interpret the IAA score? Particularly with low IAA scores, it is reasonable to ask why the scores are so low, and how this could be improved. Reasons for low IAA scores include: unclear annotation guidelines, ambiguity in the data, and having one or more unreliable annotator(s). The final annotation guidelines should be provided as supplementary materials with your final report. All annotations and output data (e.g.\ train, validation, and test outputs, possibly with confidence scores) should of course also be shared.

\section{(Overcoming) barriers to adoption}\label{sec:future}
One reason why authors may feel hesitant about providing an error analysis is that it takes up significantly more space than the inclusion of some overall performance statistics. The current page limits in our field may be too tight to include an error analysis. Relegating error analyses to the appendix does not feel right, considering the amount of work that goes into providing such an analysis. Given the effort that goes into an error analysis, authors have to make trade-offs in their time spent doing research. If papers can easily get accepted without any error analysis, it is understandable that this additional step is often avoided. How can we encourage other NLG researchers to provide more error analyses, or even just examples of errors? %

\paragraph{Improving our standards} We should adopt reporting guidelines that stress the importance of error analysis in papers reporting NLG experiments. The NLP community is already adopting such guidelines to improve the reproducibility of published work (see  \citeauthor{dodge-etal-2019-show}'s (\citeyear{dodge-etal-2019-show}) reproducibility checklist that authors for EMNLP2020 need to fill in). We should also stress the importance of error reporting in our reviewing forms; authors should be rewarded for providing insightful analyses of the outputs of their systems. One notable example here is COLING \citeyear{coling-2018-international}, which explicitly asked about error analyses in their reviewing form for NLP engineering experiments, and had a `Best Error Analysis' award.\footnote{\url{https://coling2018.org/paper-types/}}$^{,}$\footnote{\url{http://coling2018.org/index.html\%3Fp=1558.html}}

\paragraph{Making space for error analyses} We should make space for error analyses. The page limit in *ACL conferences is already expanding to incorporate ethics statements, to describe the broader impact of our research. This suggests that we have reached the limits of what fits inside standard papers, and an expansion is warranted. An alternative is to publish more journal papers, where there is more space to fit an error analysis, but then we as a community also need to encourage and increase our appreciation of journal submissions.

\paragraph{Spreading the word} Finally, we should inform others about how to carry out a proper error analysis. If this is a problem of exposure, then we should have a conversation about the importance of error reporting. This paper is an attempt to get the conversation started.

\section{Follow-up work}\label{sec:discussion}
What should you do after you have carried out an error analysis? We identify three directions for follow-up studies.

\paragraph{Errors in inputs} An additional step can be added during the identification of errors which focuses on observing the system inputs and their relation to the errors. Errors in the generated text may occur due to semantically noisy \cite{dusek-etal-2019-semantic} or incorrect system input \cite{clinciu-etal-2021-commonsense}; for instance, input data values might be inaccurate or the input might not be updated due to a recent change (e.g., new president). To pinpoint the source of the errors, we encourage authors to look at their input data jointly with the output, so that errors in inputs can be identified as such. 

\paragraph{Building new evaluation sets} Once you have identified different kinds of errors, you can try to trace the origin of those errors in your NLG model, or posit a hypothesis about what kinds of inputs cause the system to produce faulty output. But how can you tell whether the problem is really solved? Or how can you stimulate research in this direction? One solution, following \citet{mccoy-etal-2019-right}, is to construct a new evaluation set based on the (suspected) properties of the errors you have identified. Future research, knowing the scope of the problem from your error analysis, can then use this benchmark to measure progress towards a solution.

\paragraph{Scales and types of errors}
Error types and human evaluation scales are closely related. For example, if there are different kinds of grammatical errors in a text, we expect human grammaticality ratings to go down as well. But the relation between errors and human ratings is not always as transparent as with grammaticality. \Citet{van-miltenburg-etal-2020-gradations} show that different kinds of semantic errors have a different impact on the perceived overall quality of image descriptions.\footnote{Relatedly, \citet{freitag2021experts} ask annotators to rate the severity of errors in machine translation output, rather than simply marking errors.} Future research should aim to explore the connection between the two in more detail, so that there is a clearer link between different kinds of errors and different quality criteria \cite{belz-etal-2020-disentangling}.

\section{Conclusion}
Having found that NLG papers tend to underreport errors, we have motivated why authors should carry out error analyses, and provided a guide on how to carry out such analyses. We hope that this paper paves the way for more in-depth discussions of errors in NLG output.

\section*{Acknowledgements}
We would like to thank Emily Bender and the anonymous reviewers for their insightful feedback. Dimitra Gkatzia’s contribution was supported under the EPSRC projects CiViL (EP/T014598/1) and NLG for Low-resource Domains (EP/T024917/1). 
Miruna Clinciu's contribution is supported by the EPSRC Centre for Doctoral Training in Robotics and Autonomous Systems at Heriot-Watt University and the University of Edinburgh. Miruna Clinciu's PhD is funded by Schlumberger Cambridge Research Limited (EP/L016834/1, 2018-2021).
Ond\v{r}ej~Du\v{s}ek's contribution was supported by Charles University grant PRIMUS/19/SCI/10. Craig Thomson's work is supported under an EPSRC NPIF studentship grant (EP/R512412/1). Leo Leppänen's work has been supported by the European Union's Horizon 2020 research and innovation program under grant 825153 (EMBEDDIA). 

\bibliographystyle{acl_natbib}
\bibliography{anthology,acl2021}

\appendix
\section{Annotation}\label{app:annotation}
\subsection{Procedure and definitions}
We annotated all papers from INLG2010, ENLG2015, and INLG2020 in two rounds. Round 1 was an informal procedure where we generally checked whether the papers mentioned any errors at all (broadly construed, without defining the term `error'). Following this, we determined our formal annotation procedure, based on the example papers: first check if the paper is amenable. If so, check if it (a) mentions any errors in the output or (b) contains an error analysis. We used the following definitions:

\begin{description}[wide]
\item[Amenable]	A paper is amenable to an error analysis if one of its primary contributions is presenting an NLG system that produces some form of output text. So, NLG experiments are amenable to an error analysis, while survey papers are not.

\item[Error] Errors are (countable) instances of something that is wrong about the output.

\item[Error mention] An `error mention' is a reference to such an instance or a class of such instances.

\item[Error analysis] Error analyses are defined as formalised procedures through which annotators identify and categorise errors in the output.
\end{description}

\subsection{Discussion points}
The most discussion took place on the topic of amenability. Are papers that just generate prepositions \citep{muscat-belz-2015-generating} or attributes for referring expressions \citep{theune-etal-2010-cross} amenable to error analysis? And what about different versions of SimpleNLG? (E.g., \citealt{kuanzhuo-etal-2020-simplenlg}.) Although these topics feel different from, say, data-to-text systems, we believe it should be possible to carry out an error analysis in these contexts as well. In the end, amenability for us is just an artificial construct to address the (potential) criticism that we cannot just report the amount of error analyses as a proportion of all *NLG papers. As such, our definition for amenability is just a quick heuristic. Determining whether a paper really benefits from an error analysis is a more complex issue, that depends on many contextual factors.

\section{Papers containing error analyses}\label{app:summaries}
Below is a brief summary of the error analyses that we found in our annotation study.
\begin{enumerate}[wide]
\item \citet{barros-lloret-2015-input} investigate the use of different seed features for controlled neural NLG. They analysed all the outputs of their model, and categorised them based on existing lists of common grammatical errors and drafting errors.

\item \citet{akermi-etal-2020-tansformer} explore the use of pre-trained transformers for question-answering. They conducted a human evaluation study, asking 20 native speakers to indicate the presence of errors in the outputs of a French and English system. These errors were categorised as: \textit{extra
words, grammar, missing words, wrong preposition, word order}.

\item \citet{beauchemin-etal-2020-generating} aim to generate explanations of \textit{plumitifs} (dockets), based on the text of the dockets themselves. Following the identification of different errors (defined by the authors as ``the lack of realizing a specific part (accused, plaintiff or list of charges paragraphs), instead of evaluating the textual generation,'' they trace the source of the error back to either an earlier information extraction step, or to the generation procedure. 

\item \citet{kato-etal-2020-bert} present a BERT-based approach to simplify Japanese sentence-ending predicates. They took a bottom-up approach to classify the 140 cases where their model could not generate any acceptable cases. The authors then relate the error types to different stages of the generation process, and to the general architecture of their system.

\item \citet{obeid-hoque-2020-chart} present a neural NLG model for automatically providing natural language descriptions of information visualisations (i.e., charts). They manually assessed 50 output examples, and highlighted the different errors in the text. The authors find that, despite their efforts to prevent it, their model still suffers from hallucination. They identify two kinds of hallucination: either the model associates an existing value with the wrong data point, or it simply predicts an irrelevant token.
\end{enumerate}

A \textbf{notable exception} is the paper by \citet{thomson-reiter-2020-gold}, who carry out an error analysis of existing output data from three different systems. This paper was not considered amenable, because it does not present an NLG system of its own, and thus it was not included in our counts. But even if we were to count this paper among the error analyses, the trend remains the same: very few papers discuss errors in NLG output.
\end{document}